\documentclass{article}
\usepackage[utf8]{inputenc}
\usepackage{authblk}
\usepackage{biblatex}
\usepackage{graphicx, caption}
\usepackage[a4paper, total={6in, 9in}]{geometry}
\bibliography{references}

\title{}
\author[1]{Siegfried M. Ludwig}
\affil[1]{\small{MSc Artificial Intelligence, Radboud University}}

\date{August 2019}

\begin{document}

\maketitle

\begin{abstract}
Stochastic resonance describes the utility of noise in improving the detectability of weak signals in certain types of systems. It has been observed widely in natural and engineered settings, but its utility in image classification with rate-based neural networks has not been studied extensively. In this analysis a simple LSTM recurrent neural network is trained for digit recognition and classification. During the test phase, image contrast is reduced to a point where the model fails to recognize the presence of a stimulus. Controlled noise is added to partially recover classification performance. The results indicate the presence of stochastic resonance in rate-based recurrent neural networks.
\end{abstract}

\section{Introduction}
This paper investigates the occurrence of stochastic resonance (SR) in rate-based artificial neural networks (ANN), with SR describing the improved detectability of weak signals with the introduction of noise in certain types of systems. A recurrent neural network (RNN) is trained to recognize the presence of a stimulus and classify it into digits. At test time, the input intensity is reduced to a sub-threshold level and controlled noise is added to induce SR. The Python implementation is publicly available on GitHub\footnote{https://github.com/SMLudwig/lstm-stochastic-resonance}.

Considerable research has been done on the occurrence of SR in various types of natural and engineered systems (see \cite{mcdonnell2009stochastic} for a recent review) and in spiking-based neural networks (SNN). SNNs in their current state have many drawbacks however, which is the reason why rate-based models are much more popular in practice.

While benefits of general noise introduction have been studied in rate-based models, including deep learning models, those studies focused on the broader benefit of noise, particularly to make models robust against noise itself and to combat overfitting \cite{saon2019sequence}. Literature based on narrower definitions of SR relating to signal detectability is scarce for rate-based ANN and its applications in deep learning. In the following, the basic concept of more narrowly defined SR will be presented and its occurrence in natural neural systems will be introduced. Building on this base, the literature on SR in ANN will be introduced.

\subsection{Stochastic resonance}
SR is the benefit of optimal levels of noise for the detectability of weak input signals in nonlinear systems with some sort of threshold \cite{hanggi2002stochastic}. Multiple experiments have shown the benefit of optimal noise levels for detection of weak input signals in animals and humans, including the detection of sub-threshold visual stimuli (see \cite{aihara2010does} for a recent review).

\begin{figure}[thb]
    \centering
	\includegraphics[trim={0cm 0.25cm 0cm 0cm}, width=0.6\linewidth]{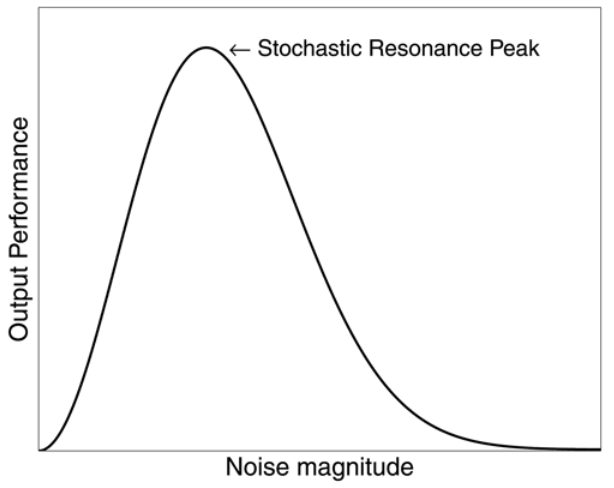}
	\captionsetup{width=.55\linewidth}
	\caption{Typical output performance versus noise level in systems capable of stochastic resonance. Adapted from \cite{mcdonnell2009stochastic}.}
	\label{fig:sr}
\end{figure}

A typical curve of the output performance at different levels of noise of a system capable of exhibiting SR is given in figure \ref{fig:sr}. Performance follows a bell-shaped curve with an optimal level of noise, below and above which performance diminishes. This behavior justifies the term stochastic resonance.

Since ANN are at least loosely based on natural neural networks, the aforementioned findings suggest that introducing noise to the input of ANN could increase signal detection performance.

\subsection{SR in artificial neural networks}
Reviewing the above mentioned definition of SR reveals the two conditions of nonlinearity and the presence of some threshold. ANN are highly nonlinear systems, a feature built in purposefully to give them universal function approximation power \cite{csaji2001approximation}. Non-linearity is usually introduced with activation functions like the sigmoid, tanh or ReLu, which are applied to the weighted and summed inputs to each neuron. While spiking neural networks clearly exhibit thresholding behavior, the situation is less clear for rate-based ANN, which by definition do not act on thresholds as straightforwardly. Nevertheless, both by introducing certain activation functions such as rectified linear units (ReLU), which is zero for negative inputs and linear for positive inputs, and by introducing a no-signal output class, threshold-like behavior can be implemented. SR could therefore be observed in certain rate-based ANN. Indeed there is theoretical mathematical work on SR in ANN, both spiking and continuous \cite{patel2008stochastic}.

Reviewing the literature reveals the occurrence of SR in Spiking neural networks (SNN) \cite{guo2009stochastic}, which are much more closely modelled on the human brain than rate-based ANN, with the latter including the deep learning models which recently have enjoyed much success. SR has also been observed in certain kinds of recurrent neural networks (RNN) \cite{krauss2018stochastic, monterola2005noise}. However, literature on more narrowly defined SR as noise-benefit in signal detection in the widely used long short-term memory (LSTM) RNN model is scarce. SR benefits for common rate-based models on common application tasks like digit classification is also scarce. Literature on SR in feed-forward ANN without recurrent connections is scarce, which is likely due to weak theoretical arguments for the occurrence of SR in those models. Since they only run inputs once through the network and lack recurrent connections, no temporal effects can emerge and the emergence of SR in the general form is therefore less conceivable.

\subsection{Research aim and relevance}
The aim of the project is to answer the question of whether and under which conditions the introduction of controlled noise can lead to SR and thereby improved signal detection in LSTM RNN. Due to the nonlinear and temporal nature of RNN and building on successes in previous literature on other kinds of ANN and RNN specifically (see introduction), SR is hypothesized to occur in LSTM RNN. Furthermore, the effect of noise on classification performance beyond mere signal detection on weak inputs will be studied as measured in the digit classification accuracy.

SR has not been studied satisfactorily in rate-based neural networks, even though they are by far the dominant network architecture, underlying deep learning systems that enjoy massive popularity. Since natural stimuli rarely resemble clear images with high contrast, robustness against weak input signals has practical utility. Potential performance enhancements and increased robustness resulting from the introduction of SR phenomena could therefore lead to many practical benefits.

\section{Method}
The model consists of a single LSTM layer with 20 hidden units and ReLU activation, followed by a dense layer with softmax activation. The ReLU activation theoretically serves as an additional thresholding function, since only values above zero do not get clipped to zero by the ReLU function. This setup roughly corresponds to a simple multilayer perceptron in the case of a sequence length of one.

The model will be trained on the MNIST digit dataset with a manually added empty stimulus class. For the empty class, black images and an additional label are added. During training, the model gets visual inputs with normal signal intensity and has to classify whether an input digit is present and if so, which digit it is. This results in a classification problem with eleven classes, ten for the input digits and an eleventh for no input. During test time, the input will be reduced to sub-threshold intensity by multiplying with a number (thresholding factor), leading the model to wrongly classify inputs as containing no signal (see figure \ref{fig:examples}).

The model is not trained on any thresholding or noise influence, since this might skew results and for practical applications it is more useful if noise can just be added during test phase, allowing the use of pre-trained models.

Following this state, controlled noise of different types and levels is introduced to the input and changes in the signal detection rate and the digit classification accuracy are observed. Uniform noise is sampled from a range between zero and a maximum noise level determined via grid search. Gaussian noise is sampled from a normal distribution with zero mean and standard deviation set by the noise level parameter during grid search. For Gaussian noise, noise values are clipped to be equal or greater to zero to avoid negative noise, which would remove parts of the stimulus even further.

The effect of different sequence lengths is studied. Theoretically longer sequences should provide the model with more information since at every step of the sequence, new noise is added to the original image. This potentially reveals different parts of the stimulus at each step.

\begin{figure}[thb]
\centering
\begin{minipage}{.475\textwidth}
    \centering
	\includegraphics[trim={0.5cm 0.5cm 0.5cm 0cm}, width=0.9\linewidth]{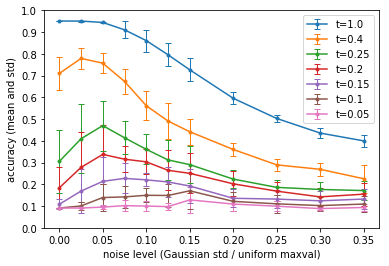}
	\caption{Uniform noise: mean and standard deviation of validation accuracy with different thresholding factors and uniform noise levels. t=1.0 corresponds to the original image.}
	\label{fig:noise_uniform}
\end{minipage}
\hfill
\begin{minipage}{.475\textwidth}
    \centering
	\includegraphics[trim={0.5cm 0.5cm 0.5cm 0cm}, width=0.9\linewidth]{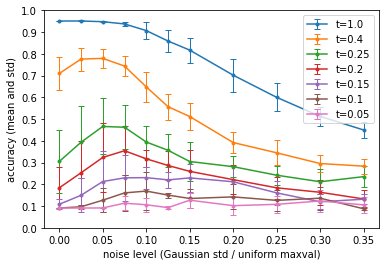}
	\caption{Gaussian noise: mean and standard deviation of validation accuracy with different thresholding factors and Gaussian noise levels. t=1.0 corresponds to the original image.}
	\label{fig:noise_gaussian}
\end{minipage}
\end{figure}

\section{Results}
Results are reported as mean and standard deviations of the model accuracy for each setup across five independently trained models. Figures \ref{fig:noise_uniform} and \ref{fig:noise_gaussian} show a clear positive effect of noise on performance under certain conditions, resembling the typical curve of stochastic resonance (see figure \ref{fig:sr}).

It can be observed that the classification of lower contrast stimuli (higher thresholding factor) benefit from higher noise levels, while higher contrast stimuli benefit from lower noise levels. Uniform (figure \ref{fig:noise_uniform}) and Gaussian noise (figure \ref{fig:noise_gaussian}) show very similar results, with the latter showing best performance for slightly higher noise levels. This difference is expected since Gaussian noise with mean at zero as in this analysis is on average smaller than uniform noise.

\begin{figure}[thb]
\centering
\begin{minipage}{.475\textwidth}
    \centering
	\includegraphics[trim={0.5cm 0.5cm 0.5cm 0cm}, width=0.9\linewidth]{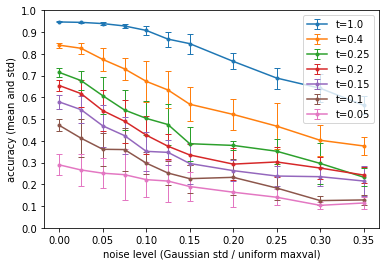}
	\caption{Without no-signal class: mean and standard deviation of validation accuracy with different thresholding factors and uniform noise levels. t=1.0 corresponds to the original image.}
	\label{fig:no_empty}
	
    \centering
	\includegraphics[trim={0.5cm 0.5cm 0.5cm -0.5cm}, width=0.9\linewidth]{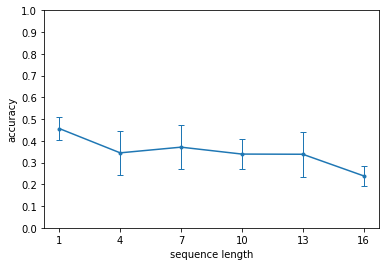}
	\caption{Mean and standard deviation of validation accuracy for different LSTM sequence lengths (t-factor=0.15, uniform noise level=0.075).}
	\label{fig:seq_len}
\end{minipage}
\hfill
\begin{minipage}{.475\textwidth}
    \centering
	\includegraphics[trim={0.5cm 0.25cm 0.5cm 0cm}, width=0.9\linewidth]{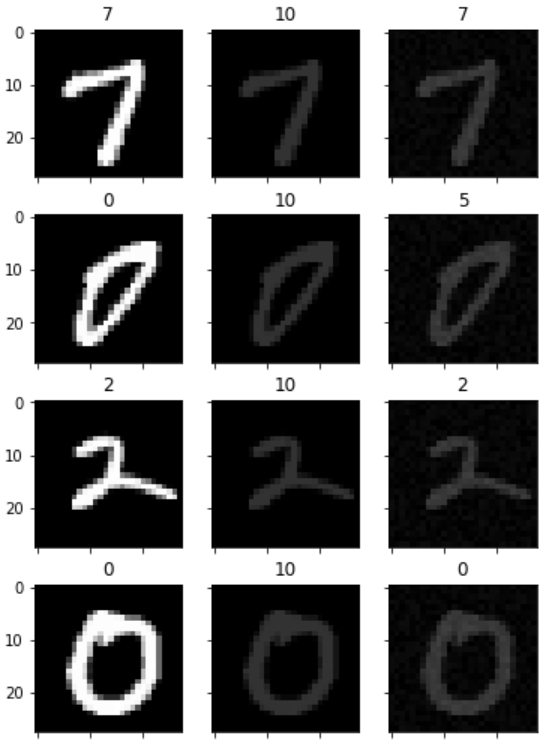}
	\caption{Examples of the effect of noise on model validation performance (t-factor=0.2, uniform noise level=0.05). Model predictions are given as numbers above each image, with 10 representing the empty class (no stimulus).}
	\label{fig:examples}
\end{minipage}
\end{figure}

Running the experiment without the added no-signal class destroys the SR effect (figure \ref{fig:no_empty}). This means that the threshold-like behavior of the no-signal class is essential to the occurrence of the SR phenomenon. ReLU activation functions do not seem to be sufficient to introduce the desired effect.

An analysis of different sequence lengths reveals almost no effect, with longer sequences even having somewhat lower performance (figure \ref{fig:seq_len}). This is likely due to the fact that the model is not trained on any noise and therefore gets identical inputs at each sequence step during training. It can not adapt to making use of the small variations across the sequence that the noise introduces. Further research could determine the utility of training a sequence model with some level of noise and applying it to sub-threshold stimuli with added noise. Reported in the figure are mean and standard deviations of the validation accuracy over five independent iterations for each sequence length.

Figure \ref{fig:examples} shows examples of the effect of noise on model performance on validation data. The first image of each row is the original image resembling the images the model is trained on. The model performance is very high on these images with accuracies around 95\%. The second image of each row is the watered down version, achieved by multiplying the image with a thresholding factor. The model usually classifies these images as containing no stimulus, which means the stimulus is below the threshold internal to the model. The last image of each row shows the watered down version with added noise. It can be seen that the model usually recognizes that some stimulus is present and to some degree accuracy also recovers, although not to the full level.

\section{Discussion}
The LSTM model trained on digit recognition with an added class of empty stimuli exhibited clear SR behavior. Adding noise to sub-threshold stimuli partially recovered classification performance, with the optimal noise level being dependent on the stimulus intensity. This shows that rate-based recurrent neural networks do exhibit SR and can be made more robust against weak stimuli by making use of the SR phenomenon.

Interestingly, a sequence length of one exhibits at least equal performance to longer sequences (figure \ref{fig:seq_len}). This means that the recurrent nature of the model studied here is not a prerequisite for the occurrence of SR. Other models, like convolutional neural networks, could therefore also exhibit SR. This is a promising direction for future research.

The dependence of the optimal noise level on stimulus intensity presents a problem in practice, since the stimulus intensity can not be controlled in the way it is in this experiment. Potential ways do deal with this problem would be to preprocess stimuli or to add noise adaptively depending on the stimulus intensity. An intersting way could be to make use of the recurrent architecture by introducing varying noise levels at each step of the sequence, which could allow the model to pick the optimal noise level itself.

The results also show that running the experiment without the added no-signal class gives better performance even though no SR effect is exhibited. On the type of controlled task the MNIST dataset represents, this is an argument against making use of noise to introduce SR. On object recognition tasks and in practical applications however, instances with absence of stimuli are to be expected and the no-signal class is a natural part of the problem formulation. Under those more practical conditions, SR effects could improve performance.

\printbibliography

\end{document}